\documentclass[preprint,12pt]{elsarticle}

\usepackage{amssymb}
\usepackage{graphicx}
\usepackage{amsmath}
\usepackage{subcaption}
\usepackage{algorithm}
\usepackage{algpseudocode}
\usepackage{multirow}
\usepackage{tikz}
\usepackage{epstopdf}
\usepackage{amsthm} 
\usepackage{bbm}
\usepackage{booktabs}       
\usepackage{amsfonts}       
\usepackage{nicefrac}       
\usepackage{microtype} 
\usepackage{newfloat}
\usepackage{listings}
\usepackage{times}  
\usepackage{helvet}  
\usepackage{courier}  
\usepackage[hyphens]{url}  
\usepackage{xcolor}         
\usepackage{natbib} 
\usepackage{caption} 
\usepackage{subcaption}
\usepackage{bbm}
\usepackage{lipsum}
\newtheorem{theorem}{Theorem}
\newtheorem{lemma}{Lemma}
\newtheorem{definition}{Definition}[section]

\newcommand*{\circled}[1]{\lower.7ex\hbox{\tikz\draw (0pt, 0pt)%
    circle (.5em) node {\makebox[1em][c]{\small #1}};}}
\usepackage{listings}
\DeclareCaptionStyle{ruled}{labelfont=normalfont,labelsep=colon,strut=off} 
\floatstyle{ruled}
\newfloat{listing}{tb}{lst}{}
\floatname{listing}{Listing}

\begin{document}

\begin{frontmatter}



\title{Lower Difficulty and Better Robustness: \\ A Bregman Divergence Perspective for Adversarial Training}


\author{Zihui Wu}

\author{Haichang Gao \corref{cor1}}
\cortext[cor1]{Corresponding author}

\author{Bingqian Zhou}

\author{Xiaoyan Guo}

\author{Shudong Zhang\blfootnote{zihui@stu.xidian.edu.cn(Z.Wu); hchgao@xidian.edu.cn(H.Gao);zbqxidian@stu.xidian.edu.cn\\(B.Zhou);xiaoyanguo@stu.xidian.edu.cn(X.Guo); sdong.zhang@outlook.com(S.Zhang)}} 

\address{School of Computer Science and Technology, Xidian University.
, Xi’an 710071, Shaanxi, China}
\newcommand\blfootnote[1]{%
\begingroup
\renewcommand\thefootnote{}\footnote{#1}%
\addtocounter{footnote}{-1}%
\endgroup
}

\begin{abstract}

In this paper, we investigate on improving the adversarial robustness obtained in adversarial training (AT) via reducing the difficulty of optimization. To better study this problem, we build a novel Bregman divergence perspective for AT, in which AT can be viewed as the sliding process of the training data points on the negative entropy curve.
Based on this perspective, we analyze the learning objectives of two typical AT methods, \emph{i.e.}, PGD-AT and TRADES, and we find that the optimization process of TRADES is easier than PGD-AT for that TRADES separates PGD-AT. In addition, we discuss the function of entropy in TRADES, and we find that models with high entropy can be better robustness learners. Inspired by the above findings, we propose two methods, \emph{i.e.}, \textbf{FAIT} and \textbf{MER}, which can both not only reduce the difficulty of optimization under the 10-step PGD adversaries, but also provide better robustness.
Our work suggests that reducing the difficulty of optimization under the 10-step PGD adversaries is a promising approach for enhancing the adversarial robustness in AT.

\end{abstract}

\begin{graphicalabstract}
\end{graphicalabstract}

\begin{highlights}
\item Research highlight 1
\item Research highlight 2
\end{highlights}


\end{frontmatter}


\section{Introduction}
\label{}

\begin{figure}[t]
    \centering
    \includegraphics[width=0.9\linewidth]{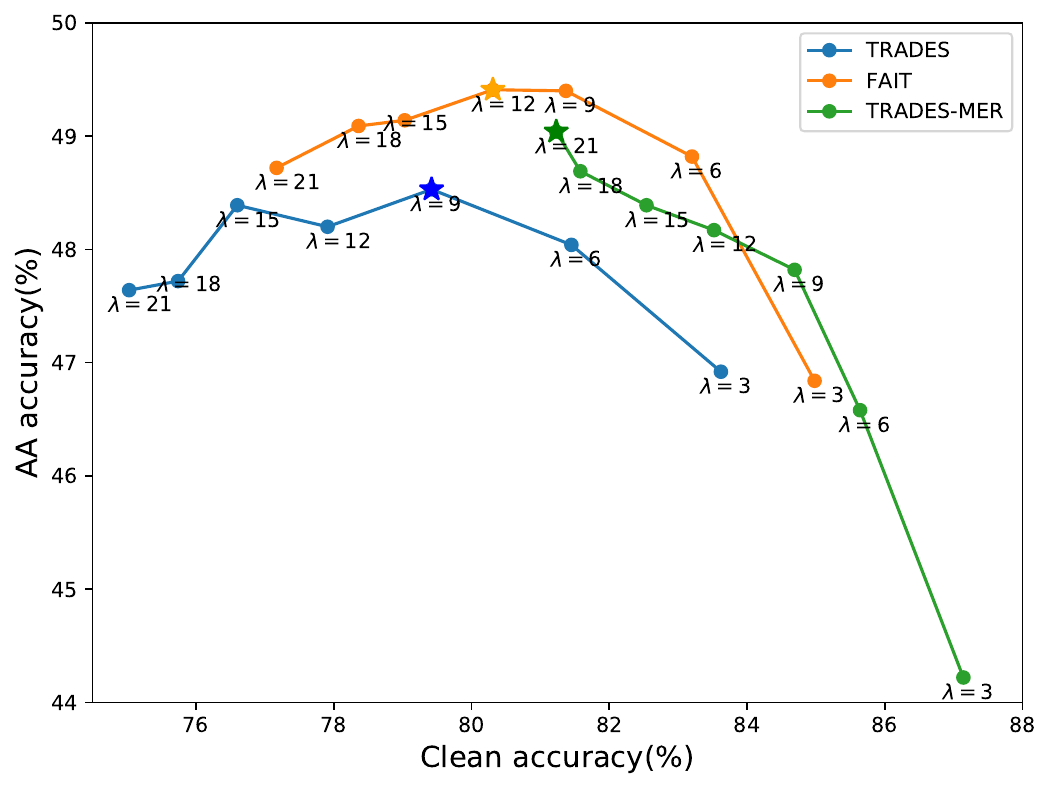}
	\caption{Both of the proposed MER and FAIT can help mitigate the robustness-accuracy tradeoff, and provide better robustness than previous baseline TRADES under the AutoAttack (AA) \cite{cite18} on CIFAR-10 with ResNet-18. $\lambda$ is a hyperparameter that balances the tradeoff, and the best robustness $\lambda$ is in the star marker.}
	\label{fig:aa}
\end{figure}

Training not only robust but also highly accurate models via \emph{adversarial training} (AT) has been found to be difficult both theoretically \cite{cite45,cite8,cite10,cite4,cite13} and empirically 
\cite{cite5,cite22,cite32}. Specifically, there exists a \emph{robustness-accuracy tradeoff} \cite{cite8} in AT 
that an increase in robustness is usually accompanied by a decrease in accuracy.
However, even though we lower the high accuracy requirement, we find that improving the robustness alone remains difficult. For instance, the previous work TRADES \cite{cite6} separates the AT learning objective into an accuracy loss $\mathcal{A}_\theta$ and a robustness loss $\mathcal{R}_\theta$ and
makes the robustness-accuracy tradeoff controllable by a hyperparameter $\lambda$. 
However, when increasing $\lambda$ for better robustness, we find the robustness of TRADES is saturated after $\lambda>9$,
but the accuracy still decreases, as the blue line in Fig.\ref{fig:aa} shows. This observation motivates our deep thinking:
why is it so difficult to improve adversarial robustness in AT?

A reasonable explanation is that, in AT, we encourage \emph{adversarial examples} to fit the 
distributions of the clean examples, \emph{e.g.,} the $\mathcal{R}_\theta$ in TRADES is
$$\mathcal{R}_\theta (x,x^\prime) = KL(p_\theta(x) \Vert p_\theta(x^\prime) ) .$$
However, the underlying distributions between adversarial examples and clean examples could be very different \cite{cite52,cite51,cite54}, in which we can even distinguish them via training a classifier \cite{cite53}.
This makes it difficult to fit adversarial examples with the distributions of clean examples.
As a result, the robustness loss $\mathcal{R}_\theta$ in TRADES is hard to optimize,
and the robustness cannot reach to the same high level as accuracy even with large $\lambda$ values.
Therefore, an assumption comes to our mind: \emph{could reduce the difficulty of optimizing the robustness loss $\mathcal{R}_\theta$ helps improve adversarial robustness?}

To better study this problem, we build a novel Bregman divergence perspective to examine AT.
This perspective can help us analyze the AT learning objective clearly. From this perspective, we analyze two typical AT methods, PGD-AT \cite{cite12} and TRADES \cite{cite6}, and we obtain interesting findings. First, we find that the separation of the loss function is beneficial, making TRADES easier to optimize than PGD-AT. This finding motivates us to propose \emph{\textbf{F}riendly \textbf{A}dversarial \textbf{I}nterpolation \textbf{T}raining} (\textbf{FAIT}), which separates $\mathcal{R}_\theta$ by adding an interpolated PGD adversary to reduce the
optimization difficulty of $\mathcal{R}_\theta$. Second, we study the function of entropy in AT, and
we find that models with higher entropy are better robustness learners. Motivated by this finding, we 
incorporate \emph{\textbf{M}aximum \textbf{E}ntropy \textbf{R}egularization} (\textbf{MER}) into AT, which is a classic regularization method for maximizing the entropy of the output distribution of DNN models. We verify that FAIT and MER can help reduce the optimization difficulty of $\mathcal{R}_\theta$ because both of them could adopt a larger $\lambda$ and have smaller robustness losses than their prototype method TRADES, and our methods also outperform TRADES in robustness, as shown in Fig. \ref{fig:aa}.
In addition,
we also conduct a comparison with other previous state of the art AT methods to show the effectiveness of our methods. 
At last, we present the scalability of the proposed methods to different model architectures and statistical distances.
In summary, the main contributions of this paper are as follows.
\begin{enumerate}
\item  We provide a novel Bregman perspective to examine AT, which can help analyze the learning objective of AT. From this perspective, we propose two guidelines for the AT learning objective design: \textbf{better to separate than to merge} and \textbf{high-entropy models are better robustness learners}.
\item  Following these two guidelines, we propose a novel robustness loss and a regularization method, \emph{i.e.,} FAIT and MER, both of which can reduce the optimization difficulty of the robustness loss $\mathcal{R}_\theta$ and effectively enhance the adversarial robustness of the resulting model.
\item Our work demonstrates that reducing the optimization difficulty of $\mathcal{R}_\theta$ under the 10-step PGD adversaries is a promising approach for enhancing robustness that can provide insights for future works to design more robust models and algorithms.
\end{enumerate}

	The rest of this paper is organized as follows. In Sec. \ref{sec:2}, we briefly review related works on AT. In Sec. \ref{sec:3}, we present the novel Bregman divergence perspective, and we provide theoretical analyses in the simple binary classification case. In Sec. \ref{sec:4}, we describe the FAIT and MER methods and present their implementations. The experimental results obtained on different datasets are provided in Sec. \ref{sec:5}. Finally, we conclude this paper in Sec. \ref{sec:6}.
\subsection{Notations}
In this paper, we use $f_\theta$ to denote a DNN model $f$ parameterized by $\theta$, and for each data point $(x,y)$ in the training set $\mathcal{D}^{N}_{train}$,
the corresponding probability output is denoted as $p_{\theta}(x) = softmax(f_\theta(x))$. We use $CE$ to denote the cross-entropy loss and $KL$ to denote the Kullback‒Leibler divergence (KL-divergence). We use $\mathcal{B}(x,\epsilon)$ to denote the norm balls centered on $x$ with with a radius of $\epsilon$,
and $\mathcal{B}(\mathcal{D},\epsilon)$ to denote the
collection of norm balls for $x\in \mathcal{D}$.

\section{Related work}
\label{sec:2}
\textbf{PGD-based AT}. 
The projected gradient descent (PGD)-based AT is currently the most effective approach 
to train robust DNN models, which aims to solve the following min-max optimization problem
\begin{equation}
    \label{inner_max}
     \mathop{\arg\min}_{\theta} \mathop{\arg\max}_{x^\prime} \mathbb{E} \{ \mathcal{L}(x,x^\prime,y;\theta) \}, \emph{s.t.}, x^\prime \in \mathcal{B}(x,\epsilon).
\end{equation}
In the inner maximization of Eq. \eqref{inner_max}, the PGD adversaries $x^\prime$ are obtained by iteratively executing
$$
x^\prime \leftarrow \Pi_{\mathcal{B}(x,\epsilon)}(x^\prime+\eta_{pgd} \cdot sign(\nabla_{x^\prime} \mathcal{L})),
$$
where $\Pi$ is the projection operator,
and the loss function $\mathcal{L}$ may be different in various works \cite{cite12,cite6,cite3}.

\textbf{PGD-AT } \cite{cite12}.
\citeauthor{cite12} used the cross-entropy loss as the loss function $\mathcal{L}$ in Eq. \eqref{inner_max} and
first incorporated the 10-step PGD adversary into AT to solve the outer minimization problem:
\begin{equation}
    \label{eq:pgd-at}
   \mathop{\arg\min}_{\theta} \mathbb{E} \{ CE(p_{\theta}(x^\prime) ,y) \},
\end{equation}
which is known as the PGD-AT approach. 

\textbf{TRADES} \cite{cite6}. The learning objective of TRADES, defined by Eq. \eqref{trades_obj_a},
separates the PGD-AT learning objective in Eq. \eqref{eq:pgd-at} into two parts: an accuracy loss $\mathcal{A}_{\theta}$ and a robustness loss $\mathcal{R}_{\theta}$, and $\lambda$ is a hyperparameter that balances the robustness-accuracy tradeoff.
In addition, TRADES uses the KL-divergence as the loss $\mathcal{L}$ in the inner maximization of Eq. \eqref{inner_max} rather than the cross-entropy loss in PGD-AT.
\begin{equation}
    \label{trades_obj_a}
        \mathop{\arg\min}_{\theta} \mathbb{E} \{ \underbrace{CE(p_{\theta}(x),y) }_{\mathcal{A}_{\theta}}+  \lambda \underbrace{  \cdot KL(p_{\theta}(x) \Vert p_{\theta}(x^\prime))}_{\mathcal{R}_{\theta}} \}.
\end{equation}

\textbf{Mitigating the robustness-accuracy tradeoff}.
One of the biggest problems in AT is the robustness-accuracy tradeoff \cite{cite5,cite8,cite13,cite23}.
To mitigate this tradeoff, a myriad of strategies have been proposed, including increasing the model capacity \cite{cite9}, performing dropout \cite{cite13}, increasing model smoothness \cite{cite15,cite14}, exploiting extra data \cite{cite22,cite32,cite10}, reducing the excessive margins \cite{cite19}, and refining the loss function \cite{cite3}.

\textbf{Reducing the optimization difficulty of AT}.
Among these many methods, reducing the optimization difficulty of AT is an intriguing
heuristic idea that motivates thinking about what the most important learning objective in AT is.
Along this line of thinking,\citet{cite2} proposed the FAT. Rather than conducting training with the most adversarial data, FAT uses the friendlier early-stopped PGD adversaries that can just make the model result in misclassification. 
In addition, some works have also used weaker FGSM adversaries \cite{DBLP:conf/iclr/WongRK20, DBLP:conf/aaai/ShafahiN0DDG20, NEURIPS2020_b8ce4776, DBLP:journals/corr/abs-2006-03089, DBLP:journals/corr/abs-2110-05007}. These methods have certain effects on mitigating the optimization difficulty; however, generalizing to unseen data is much harder, which has limited the robustness
of this type of approach.
In addition to using weaker adversaries, methods that treat data differently are available; these techniques reduce the optimization weights given to less important data, and examples include MART \cite{cite24}, MMA \cite{cite25} and GAIRAT \cite{cite21}. Such methods are effective against PGD adversaries, but tend to perform poorly against stronger attacks, \emph{e.g.}, the AA \cite{cite18}. To avoid the problems existing in previous works, in this paper, we reduce the difficulty of optimization under the 10-step PGD adversaries and we do not reduce the optimization weight for any data (actually, we adopt a larger $\lambda$ instead). As a result, our methods not only show better robustness in PGD attacks, but also in the stronger AA.

\section{A Bregman divergence perspective for AT}
\label{sec:3}
\subsection{Relationship between AT and Bregman divergence}

\subsubsection{KL-divergence equivalent form.}
Because of the entropy of a label $y$ is a constant value, for the learning objectives of PGD-AT and TRADES, we have the equivalent forms that Eq. \eqref{eq:pgd-at} $\sim$ Eq. \eqref{eq:pgd-at-kl} and Eq. \eqref{trades_obj_a} $\sim$ Eq. \eqref{trades_obj_b}, respectively.
\begin{equation}
\label{eq:pgd-at-kl}
    \mathop{\arg\min}_{\theta} \mathbb{E} \{ KL(y \Vert p_{\theta}(x^\prime)  ) \}.
\end{equation}
\begin{equation}
    \label{trades_obj_b}
        \mathop{\arg\min}_{\theta} \mathbb{E} \{ KL(y \Vert p_{\theta}(x)  )  + \lambda \cdot KL(p_{\theta}(x) \Vert p_{\theta}(x^\prime)) \}. 
\end{equation}

\subsubsection{Bregman divergence.}

Bregman divergence \cite{bregman1967relaxation} is a widely studied statistical distance in machine learning. Let $\psi: \Omega \rightarrow \mathbb{R} $  be a function that is: a) strictly convex, b) continuously
differentiable, c) defined on a closed convex set $\Omega$. Then, the Bregman divergence is defined as:
\begin{equation}
    \label{eq:bregman_div}
    \Delta_{\psi}(p,q)=\psi(p)-\psi(q)-\langle\nabla \psi(q), p-q\rangle,  \forall p, q \in \Omega,
\end{equation}
    which is the difference between the value of $\psi$ at $p$ and the first-order Taylor expansion of $\psi$ around $q$ evaluated at point $p$. Specially, when $\psi$ is the negative entropy function $-H$,
$$
  \psi(p) = -H(p) = \sum p_{i} \cdot log(p_{i}),
$$
the Bregman divergence degrades to the KL-divergence:
\begin{equation*}
\begin{aligned}
    KL(p \Vert q) &= \Delta_{-H}(p,q) \\
                  &= -H(p)+H(q)-\langle\nabla-H(q), p-q\rangle .
\end{aligned}
\end{equation*}

Because of the learning objective of PGD-AT and TRADES can be expressed in the KL-divergence equivalent form, as shown in Eq. \eqref{eq:pgd-at-kl} and Eq. \eqref{trades_obj_b}, and KL-divergence is one of the special cases of the Bregman divergence,
the learning objective of PGD-AT and TRADES can actually be viewed as the minimization of the Bregman divergence. Inspired by this finding,
we form a novel Bregman divergence perspective to look at AT; that is, we regard the training process of AT as both clean data points $x$ and adversarial data points $x^\prime$ sliding on the curve of the function $\psi$ (in PGD-AT and TRADES, $\psi$ is $-H$), and for PGD-AT, the target is to reduce the difference between $-H(y)$ and the first-order Taylor expansion of $-H(p_\theta(x^\prime))$ at $y$. For TRADES, the robustness loss term $\mathcal{R}_\theta$ is the difference between $-H(p_\theta(x))$ and the first-order Taylor expansion of $-H(p_\theta(x^\prime))$ at $p_\theta(x)$, and the accuracy loss term $\mathcal{A}_\theta$ is the difference between the value of $-H(y)$ and the first order Taylor expansion of $-H(p_\theta(x))$ at $y$.

\subsection{Binary classification analyses }
\begin{figure}[ht]
	\begin{subfigure}{0.49\linewidth}
	    \centering
	    \includegraphics[width=1\linewidth]{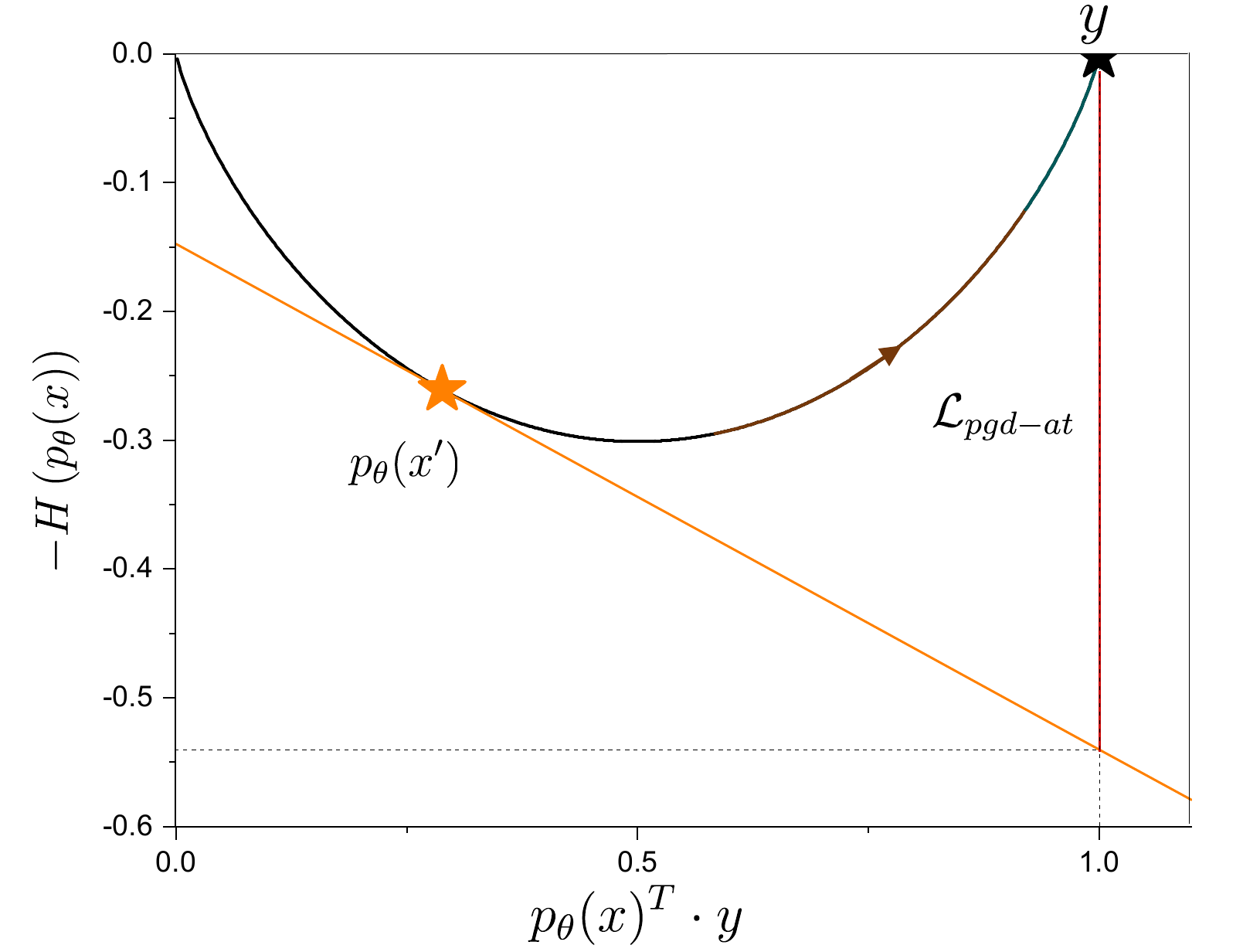}
	    \caption{PGD-AT}
	    \label{fig:pgd-at}
	\end{subfigure}
	\begin{subfigure}{0.49\linewidth}
	    \centering
	    \includegraphics[width=1\linewidth]{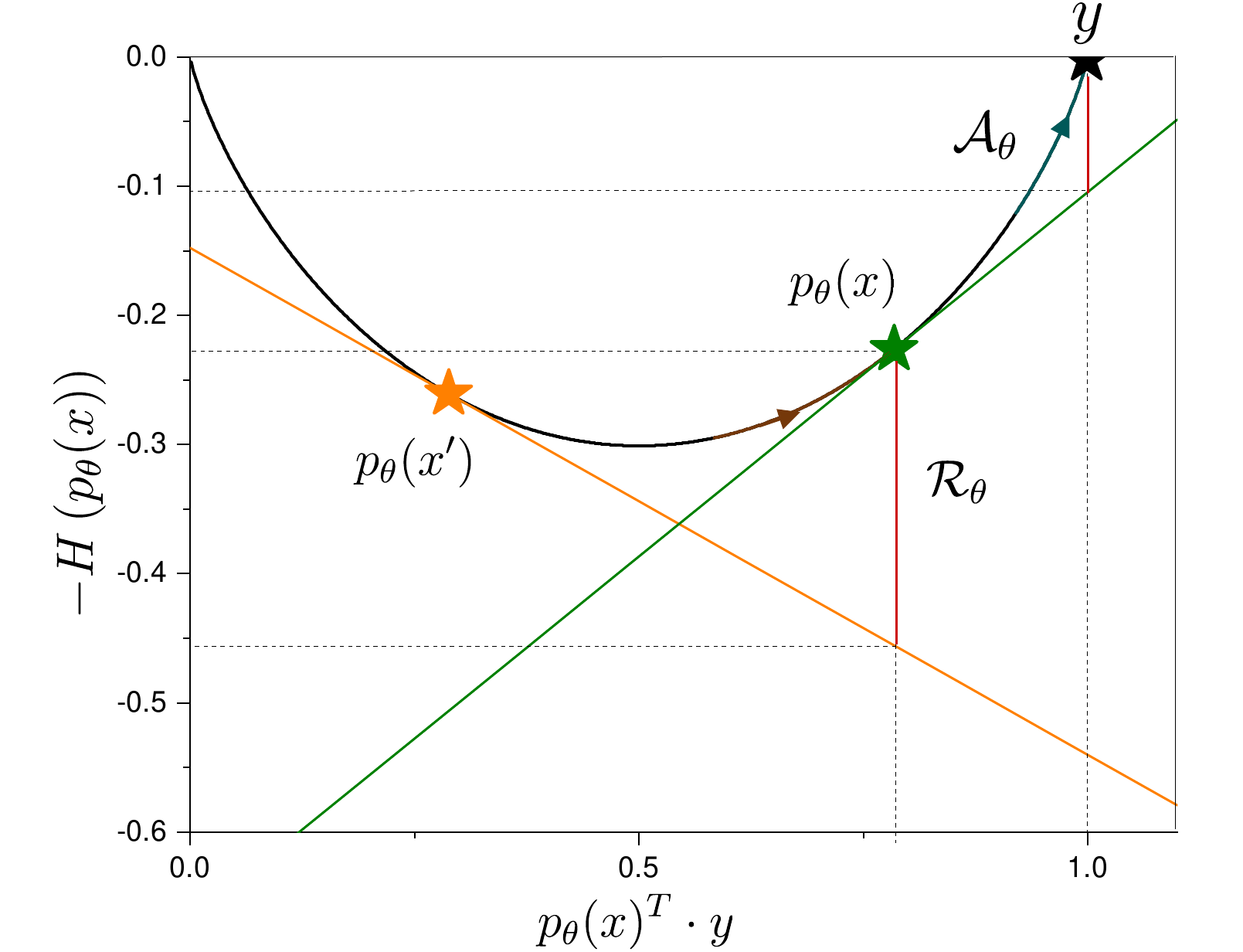}
	    \caption{TRADES}
	    \label{fig:trades}
	\end{subfigure}
\caption{Illustrations of the Bregman divergence perspective of PGD-AT and TRADES.}
\label{fig:bregman_fig}
\end{figure}

To simply explain our perspective, we show the illustration in the case of binary classification in Fig. \ref{fig:bregman_fig}. In this case, the label $y$ is a collection of 2-D one-hot vectors $[0,1]^T$ and $[1,0]^T$, $p_{\theta}(x)$ is a 2-D probability distribution and $p_{\theta}(x)^T \cdot y$ is the projection of $p_{\theta}(x)$ in the $y$ direction.
We plot the illustration of PGD-AT in Fig. \ref{fig:pgd-at}, and we can intuitively see the loss term $\mathcal{L}_{pgd-at}$ (Eq. \ref{loss:pgd-at}),  as shown by the red line, which is the difference between $-H(y)$ and the first order Taylor expansion of $-H(p_\theta(x^\prime))$ at $y$, as mentioned above.
\begin{equation}
  \label{loss:pgd-at}
  \mathcal{L}_{pgd-at} = KL(y \Vert p_{\theta}(x^\prime)) = \Delta_{-H}(y,p_{\theta}(x^\prime)).
\end{equation}
We also intuitively show the loss term of TRADES (Eq. \ref{loss:trades}) in Fig. \ref{fig:trades}.
\begin{equation}
    \label{loss:trades}
    \begin{aligned}
        \mathcal{L}_{trades} &= KL(y \Vert p_{\theta}(x))  + \lambda \cdot KL(p_{\theta}(x) \Vert p_{\theta}(x^\prime)) \\
                             &= \Delta_{-H}(y,p_{\theta}(x)) + \lambda \cdot \Delta_{-H}(p_{\theta}(x),p_{\theta}(x^\prime)) .
    \end{aligned}
\end{equation}
From this perspective,
we will then conduct the theoretical analyses in the simple binary classification case.

\subsubsection{Guideline 1: It is better to separate than to merge.}
\begin{lemma}
\label{lemma:1}
Given points $x_1$, $x_2$ and $x_*$, if $\exists \alpha \in [0,1]$ such that $p_\theta(x_*) = (1-\alpha) p_\theta(x_1) + \alpha p_\theta(x_2)  $, then the following inequality holds true:
\begin{equation*}
    \begin{aligned}
    KL(p_\theta(x_2) \Vert p_\theta(x_1)) \geq KL(p_\theta(x_2) \Vert p_\theta(x_*)) \\ 
    + KL(p_\theta(x_*) \Vert p_\theta(x_1)).
    \end{aligned}
\end{equation*}
\end{lemma}

We leave the proof of Lemma \ref{lemma:1} to the supplementary materials.
In the binary classification case, we actually have $p_\theta(x) = \alpha y + (1-\alpha) p_\theta(x^\prime), \alpha \in [0,1]$,
because $x^\prime$ is harder to classify than $x$.
According lemma 1, we can thus deduce that when $\lambda=1$ in Eq. \eqref{trades_obj_a}, the loss term $\mathcal{L}_{trades}$ is lower than $\mathcal{L}_{pgd-at}$:
\begin{equation}
    \label{eq:trades_pgdat}
    \mathcal{L}_{pgd-at} \geq \mathcal{L}_{trades} = \mathcal{A}_\theta + \mathcal{R}_\theta,
\end{equation}
as intuitively shown in Fig. \ref{fig:pgd-at} and Fig. \ref{fig:trades}. 
Eq. \eqref{eq:trades_pgdat} indicates that the separation of the learning objective does not only makes TRADES able to balance the robustness-accuracy tradeoff but also 
reduces its optimization difficulty compared to that of PGD-AT.
As a result, TRADES can adopt a larger $\lambda > 1$ and attain better robustness than PGD-AT when $\lambda$ increases to the best robustness-accuracy tradeoff value. Nevertheless, the optimization of the robustness loss $\mathcal{R}_\theta$ remains difficult. Therefore, we think:
\begin{center}
    \emph{As TRADES is the separation of PGD-AT which reduces the optimization difficulty,
    can we separate the $\mathcal{R}_\theta$ again to make TRADES easier to train?}
\end{center}
Motivated by this idea,
we proposed the FAIT, which separates $\mathcal{R}_\theta$ into two smaller units by introducing an interpolated PGD adversary. We will introduce this method in more detail in Sec. \ref{sec:4}, but before we do, let us introduce our other interesting finding.

\subsubsection{Guideline 2: High-entropy models are better robustness learners.}

As discussed, from the Bregman divergence perspective of AT, the training data slide on the negative \underline{entropy} curve. This motivates us to study the function of entropy in AT. Specifically, we aim to answer the following question:
\begin{center}
    \emph{In AT, is a model with higher entropy better, \\ or is a model with lower entropy better?}
\end{center}
To study this question, 
comparing the entropy values of different models is necessary, and we first give the following definition:
\begin{definition}[Entropy upper bound]
\label{define:1}
 Given two models $f_{\theta_{1}}$ and $f_{\theta_2}$, $\forall \tilde{x} \in \mathcal{B}(\mathcal{D},\epsilon)$, 
if the entropy of $\tilde{x}$ satisfies $ H(p_{\theta_{1}}(\tilde{x})) \leq H(p_{\theta_{2}}(\tilde{x}))  $, then we 
call model $f_{\theta_{2}}$ the entropy upper bound of model $f_{\theta_{1}}$ at a radius of $\epsilon$; this is denoted as 
$ \mathcal{H}(f_{\theta_{1}},\epsilon) \leq \mathcal{H}(f_{\theta_{2}},\epsilon).$
\end{definition}

When we have two models $f_{\theta_{1}}$ and $f_{\theta_{2}}$, however, it is not sufficient to analyze their optimization difficulty levels when $f_{\theta_{1}}$ and $f_{\theta_{2}}$ only satisfy $ \mathcal{H}(f_{\theta_{1}},\epsilon) \leq \mathcal{H}(f_{\theta_{2}},\epsilon)$. For example,
given an initial model with large entropy and a well-trained model with small entropy, 
it is inappropriate to compare the optimization difficulty levels of these two models because training the initial model is obviously much easier. That is, the convergence degrees of the two models should be similar for
a fair comparison.  
Therefore, we give the following definition to ensure that the two models have the similar convergence degrees in the adversarial context.

\begin{definition}[Identical adv-convergence]
$\forall \tilde{x} \in \mathcal{B}(\mathcal{D},\epsilon)$, if model 
$f_{\theta_{1}}$ and $f_{\theta_{2}}$ satisfy:
$$
    \mathop{\arg\max}_{i} p_{\theta_1}(\tilde{x})_i = \mathop{\arg\max}_{i} p_{\theta_2}(\tilde{x})_i
$$
then $f_{\theta_{1}}$ has identical adv-convergence with $f_{\theta_{2}}$, and this is denoted as 
$\mathbb{I}(f_{\theta_{1}},\epsilon) = \mathbb{I}(f_{\theta_{2}},\epsilon).$ 
\end{definition}

When $\mathbb{I}(f_{\theta_{1}},\epsilon) = \mathbb{I}(f_{\theta_{2}},\epsilon),$  it is easy to infer that $f_{\theta_{1}}$ and $f_{\theta_{2}}$ have the same the accuracy and robustness. That is, $f_{\theta_{1}}$ and $f_{\theta_{2}}$ will maintain the same robustness-accuracy tradeoff without the need to maintain the same entropy, and this property is useful for the analysis process.

In the binary classification case,
there are three conditions regarding the clean output distribution $p_{\theta_{1}}(x)$ and the adversarial output
distribution $p_{\theta_{1}}(x^\prime)$ in  
model $f_{\theta_{1}}$: 
$ \mathcal{C}.1$: $p_{\theta_{1}}(x)^T \cdot y > \frac{1}{2}, p_{\theta_{1}}(x^\prime)^T \cdot y  \leq \frac{1}{2} $;
$ \mathcal{C}.2$: $p_{\theta_{1}}(x)^T \cdot y \leq \frac{1}{2}, p_{\theta_{1}}(x^\prime)^T \cdot y \leq \frac{1}{2} $;
$ \mathcal{C}.3$: $p_{\theta_{1}}(x)^T \cdot y > \frac{1}{2}, p_{\theta_{1}}(x^\prime)^T \cdot y > \frac{1}{2} $.
Given a model $f_{\theta_{2}}$ that satisfies $ \mathcal{H}(f_{\theta_{1}},\epsilon) \leq \mathcal{H}(f_{\theta_{2}},\epsilon)$ and $\mathbb{I}(f_{\theta_{1}},\epsilon) = \mathbb{I}(f_{\theta_{2}},\epsilon)$,
we have the following two theorems for these different conditions.

\begin{theorem}[$ \mathcal{C}.1$]
\label{theory:1} Given $ \mathcal{H}(f_{\theta_{1}},\epsilon) \leq \mathcal{H}(f_{\theta_{2}},\epsilon)$ and $\mathbb{I}(f_{\theta_{1}},\epsilon) = \mathbb{I}(f_{\theta_{2}},\epsilon)$,
$\forall x \in \mathcal{D}$, if $\mathcal{C}.1$ holds true, then we have $\mathcal{R}_{\theta_{1}}(x,x^\prime) \geq \mathcal{R}_{\theta_{2}}(x,x^\prime) $.
\end{theorem}

\begin{theorem}[$ \mathcal{C}.2 , \mathcal{C}.3$]
\label{theory:2}
Define the difference between the clean and the adversarial probability distribution as $d_{\theta}(x,x^\prime) = p_{\theta}(x) - p_{\theta}(x^\prime) $ 
. Given $ \mathcal{H}(f_{\theta_{1}},\epsilon) \leq \mathcal{H}(f_{\theta_{2}},\epsilon)$ and $\mathbb{I}(f_{\theta_{1}},\epsilon) = \mathbb{I}(f_{\theta_{2}},\epsilon)$,
$\forall x \in \mathcal{D}$,
if $\mathcal{C}.2$ or 
$\mathcal{C}.3$ holds true,
let $d_{\theta_{1}}(x,x^\prime) = d_{\theta_{2}}(x,x^\prime)$, we also have 
$\mathcal{R}_{\theta_{1}}(x,x^\prime) \geq \mathcal{R}_{\theta_{2}}(x,x^\prime) $.
\end{theorem}

\begin{figure*}[ht]
    \begin{subfigure}{0.49\linewidth}
        \centering
        \includegraphics[scale=0.25]{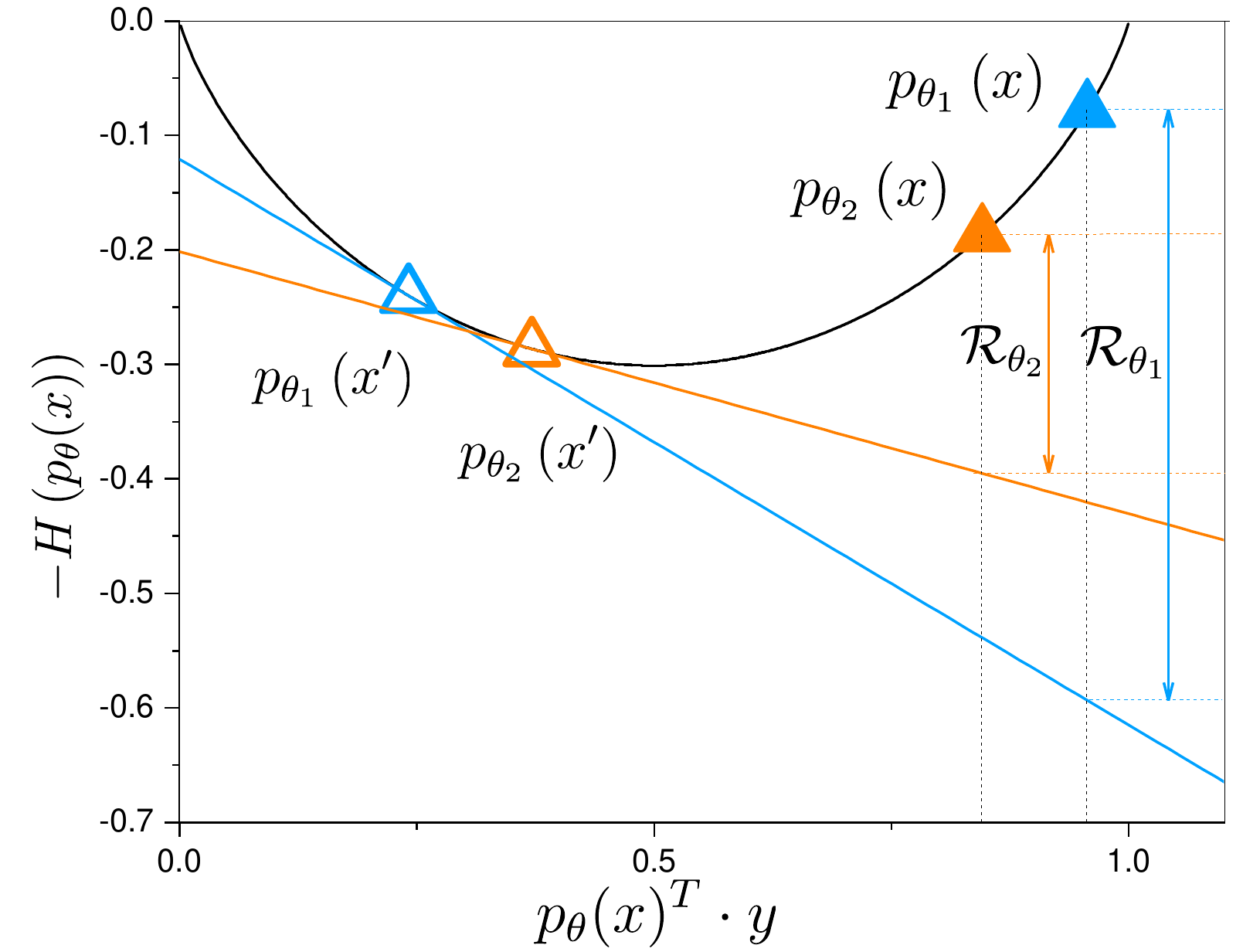}
        \caption{$\mathcal{C}.1$}
        \label{theory2:1}
    \end{subfigure}
    \begin{subfigure}{0.49\linewidth}
        \centering
        \includegraphics[scale=0.25]{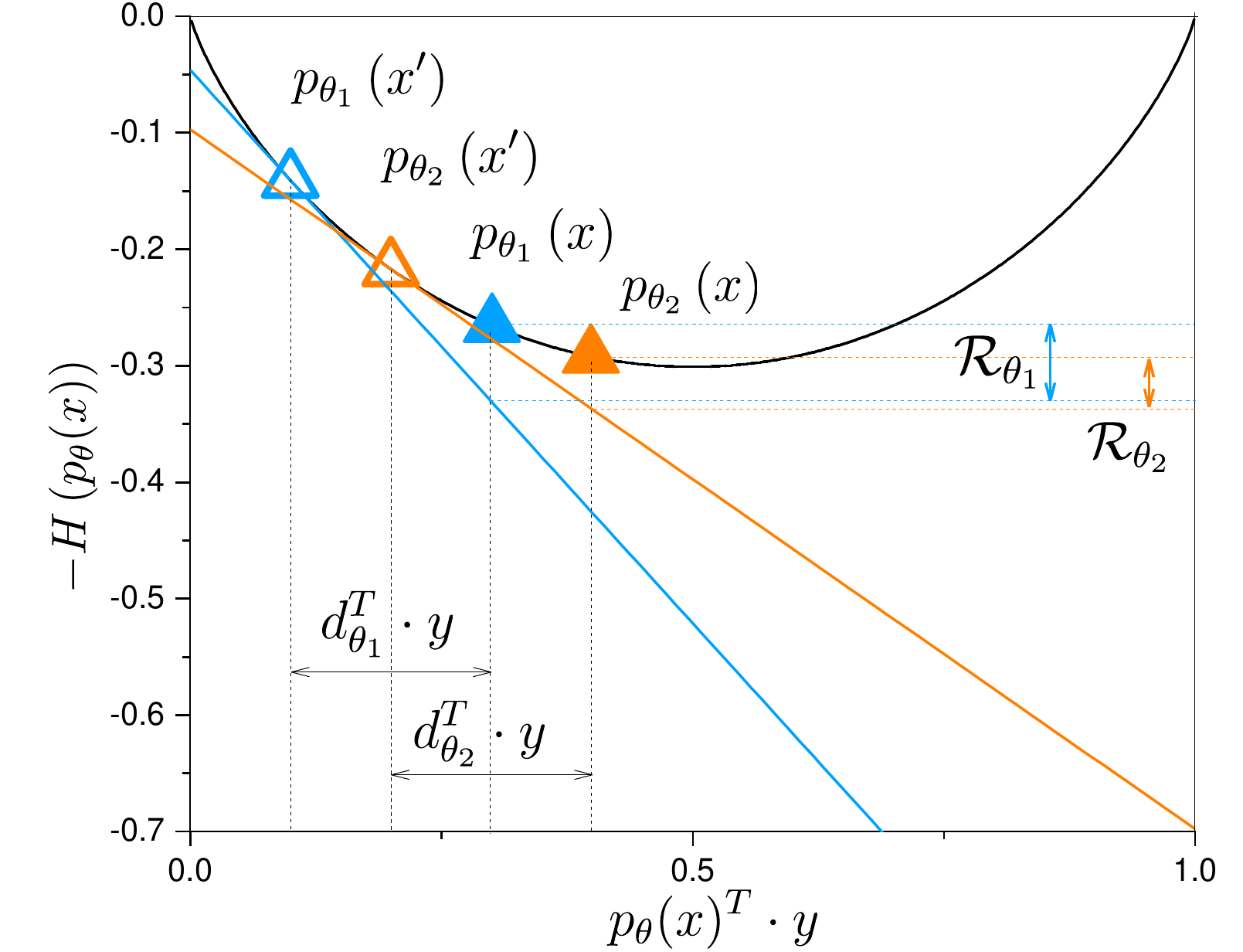}
        \caption{$\mathcal{C}.2$}
        \label{theory2:2}
    \end{subfigure}
    \begin{subfigure}{0.49\linewidth}
        \centering
        \includegraphics[scale=0.25]{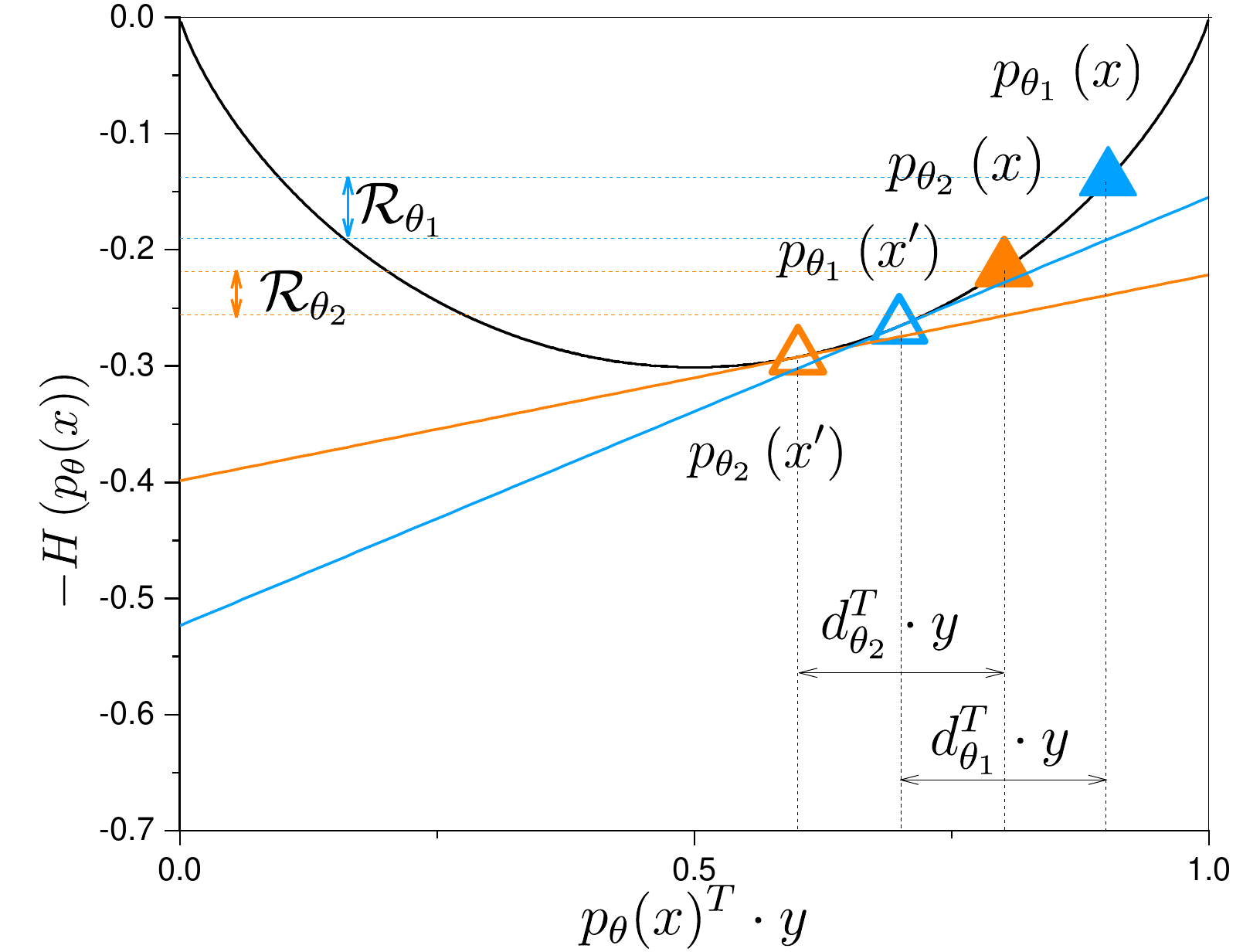}
        \caption{$\mathcal{C}.3$}
        \label{theory2:3}
    \end{subfigure}
    \caption{Illustrations of $R_\theta$ in models $f_{\theta_{1}}$ and $f_{\theta_{2}}$ when $ \mathcal{H}(f_{\theta_{1}},\epsilon) \leq \mathcal{H}(f_{\theta_{2}},\epsilon)$ at the three different conditions.}
    \label{fig:rrr}
\end{figure*}

\subsubsection{Remark 1.} Proofs are provided in the supplementary materials.
Theorem \ref{theory:1} and Theorem \ref{theory:2} tell us the following.
\begin{center}
    \emph{When two models keep the same robustness-accuracy tradeoff, the robustness loss $\mathcal{R}_{\theta}$ of the model with higher entropy is easier to optimize.}
\end{center}
We provide the corresponding illustrations of the 
three conditions in Fig. \ref{fig:rrr} for better understanding.
Based on this analysis, we introduce the MER strategy to maximize the entropy of the robust model, and we find that MER can effectively reduce the difficulty of optimizing $\mathcal{R}_\theta$.

\section{Method}
\label{sec:4}
Based on the above analyses,
in this section, we introduce two methods for mitigating the optimization difficulty of $\mathcal{R}_\theta$.
The first is FAIT, which incorporate an interpolated PGD adversary into the training process. The second is MER, which maximizes the entropy of the output distribution.

\subsection{FAIT}
To further reduce the training difficulty of TRADES, we propose 
FAIT.
FAIT separates $\mathcal{R}_\theta$ in TRADES by adding a new interpolation data point $x^*$, and replacing $\mathcal{R}_\theta$ with $\mathcal{R}^{\prime}_\theta$:
$$
    \mathcal{R}^{\prime}_\theta(x,x^*,x^\prime) = KL(p_\theta(x) \Vert p_\theta(x^*)) + KL(p_\theta(x^*) \Vert p_\theta(x^\prime)).
$$
There are various choices of $x^*$ only if that $x^*$ is more adversarial than $x$ and less adversarial than $x^\prime$.
To avoid introducing extra computational overhead for generating $x^*$, we sample $x^*$ from the PGD iteration process with a fixed interpolation number $I (0< I < K)$, where $K$ is the number of PGD iterations. 
In the inner maximization, we keep using the KL-divergence as the loss function to retain the same 10-step PGD adversary of TRADES.
In Algorithm \ref{alg:fait},
the pseudocode of FAIT is displayed for a more detailed understanding.

\begin{algorithm}[ht]
    \caption{Friendly Adversarial Interpolation Training}
    \label{alg:fait}
    \textbf{Input:} Training dataset $D^N_{train}$
    
    \textbf{Parameter:} Batch size $m$; learning rate 
    $\eta_{lr}$;
    PGD step size $\eta_{pgd}$; number of PGD iterations $K$; perturbation size $\epsilon$; PGD interpolation number $I$;
    \begin{algorithmic}[1]
       \State Randomly initialize the network parameters $\theta$ 
       \State Set $\mathcal{L}_{fait}(x,x^*,x^\prime,y;\theta) = CE(p_\theta(x),y) + \lambda \cdot \mathcal{R}^{\prime}_\theta(x,x^*,x^\prime) $
       \Repeat 
       \State Sample a mini-batch $\{(x_{i},y_{i})\}_{i=1}^{m}$ from $D_{train}^{N}$
       \For{$i=1,2,...,m$}
       \State $x_i^{\prime} \leftarrow x_i + 0.001 \cdot \mathcal{N}(0, I) $
            \For{$k=1,2,...K$}
                \State  $x^\prime_i \leftarrow \Pi_{\mathcal{B}(x_i,\epsilon)}(x_i^\prime+\eta_{pgd} \cdot sign(\nabla_{x_i^\prime}
                KL(p_\theta(x_i),p_\theta(x_i^\prime)))$
                \Comment{Generate $x^\prime$}
                \If{$k=I$}
                    \State $x_i^* \leftarrow x_i^\prime$
                    \Comment{Sample $x^*$}
                \EndIf
            \EndFor
       \EndFor
        \State $g_{fait} = \frac{1}{m} \sum_{i=1}^{m} \nabla_\theta \mathcal{L}_{fait} (x_i,x_i^*,x_i^\prime,y;\theta) $
        \State $\theta \leftarrow \theta - \eta_{lr} \cdot g_{fait} $
       \Until training completed
    \end{algorithmic}
\end{algorithm}

\textbf{Connection with FAT}.  Both FAIT and FAT \cite{cite2} introduced weaker PGD adversaries into AT, with the aim of reducing the difficulty of optimization. However, FAT discards the 10-step PGD adversary. Because generalizing to the unseen data is much harder, FAT is limited in its ability to achieve better robustness. Different from FAT, FAIT still employs the 10-step PGD adversary, as we shall see in Table \ref{tab:1}, FAIT can thus provide better robustness than FAT.

\subsection{MER}
The MER strategy has been widely studied in many areas of machine learning \cite{cite34,cite35,cite33,cite41}. Nevertheless, to the best of our knowledge, the effectiveness of MER in AT has not been investigated. The idea of MER is quite simple; by adding a negative entropy term in the original learning objective, the objective of MER is defined as:
$$
    \mathop{\arg\min}_{\theta} \mathbb{E} \{ \mathcal{L}(x,y;\theta) - \beta \cdot H(p_{\theta}(x))\} .
$$
However, in AT, the implementation of MER can be slightly more complicated.
In addition to maximizing the entropy of the clean output distribution $p_\theta(x)$, maximizing the entropy of the adversarial output distribution $p_\theta(x^\prime)$ is also optional. Therefore, when adding the MER into TRADES, we have the following objective:
\begin{equation}
    \begin{aligned}
    \mathop{\arg\min}_{\theta} \mathbb{E} \{ CE(p_{\theta}(x),y) + \lambda \cdot KL(p_{\theta}(x) \Vert p_{\theta}(x^\prime)) \\
              - (\beta_{cle} \cdot H(p_{\theta}(x)) + \beta_{adv} \cdot H(p_{\theta}(x^\prime))  ) \} .
    \end{aligned}
\end{equation}
We refer to this new learning objective as the TRADES-MER.

\textbf{MER is compatible with FAIT}. 
According to Lemma \ref{lemma:1}, $ \mathcal{R}_\theta$ is an upper bound of $\mathcal{R}^{\prime}_\theta$.
As discussed, $\mathcal{R}_\theta$ is easier to optimize in a model with higher entropy; 
therefore, MER can help reduce the upper bound of the $\mathcal{R}^{\prime}_\theta$, 
which makes $\mathcal{R}^{\prime}_\theta$ easier to optimize. Thus, MER should be compatible with FAIT.
Experimental results have demonstrated this point of view. As we shall see in Table \ref{tab:2},
FAIT-MER has better robustness than both FAIT and TRADES-MER.

\section{Experiments}
\label{sec:5}
\textbf{Training settings.} In the basic settings, we apply ResNet-18 \cite{cite30} as the model architecture, but we also provide results of  other architectures in Table \ref{tab:3}.
To generate the PGD adversaries, we set the step size $\eta_{pgd}= 2/255$ and the perturbation size $\epsilon=8/255$ under the $\ell_{\infty}$ norm, and the number of iterations is set to $K=10$. During the training process, we use the SGD optimizer with a weight decay of $5 \times 10^{-4}$ and momentum 0.9. We use a large batch size $m=512$ to speed up the training. We train models for 100 epochs, and the initial learning rate is 0.4 and decays by a factor of 0.1 at epochs 75 and 90. In addition, we introduce an extra 5 epochs to gradually warm up \cite{cite46} the model at the beginning to alleviate the performance degeneration caused by the large batch size.

\textbf{Robustness estimation.} 
We evaluate the robustness of the model by using PGD attacks and AA \cite{cite18}. Among them, the PGD attacks is less computationally expensive; thus, we evaluate the PGD robustness after per epoch training and record the epoch with the best robustness for further estimation. AA is a more advanced attack to verify the robustness via an ensemble of four diverse parameter-free attacks including three white-box attacks: APGD-CE \cite{cite18}, APGD-DLR \cite{cite18}, FAB \cite{cite42} and a black-box attack: Square Attack \cite{cite43}, which has been consistently shown to provide reliable robustness estimates. 

\textbf{Reproducibility.}
We report the results average over 3 runs obtained on a machine with 4 RTX 2080 Ti GPUs, and the code is provided in the supplementary materials and will be made public after the review process is completed.

\subsection{Do FAIT and MER reduce the difficulty of optimization?}
\begin{figure}[ht]
    \centering
    \includegraphics[width=0.8\linewidth]{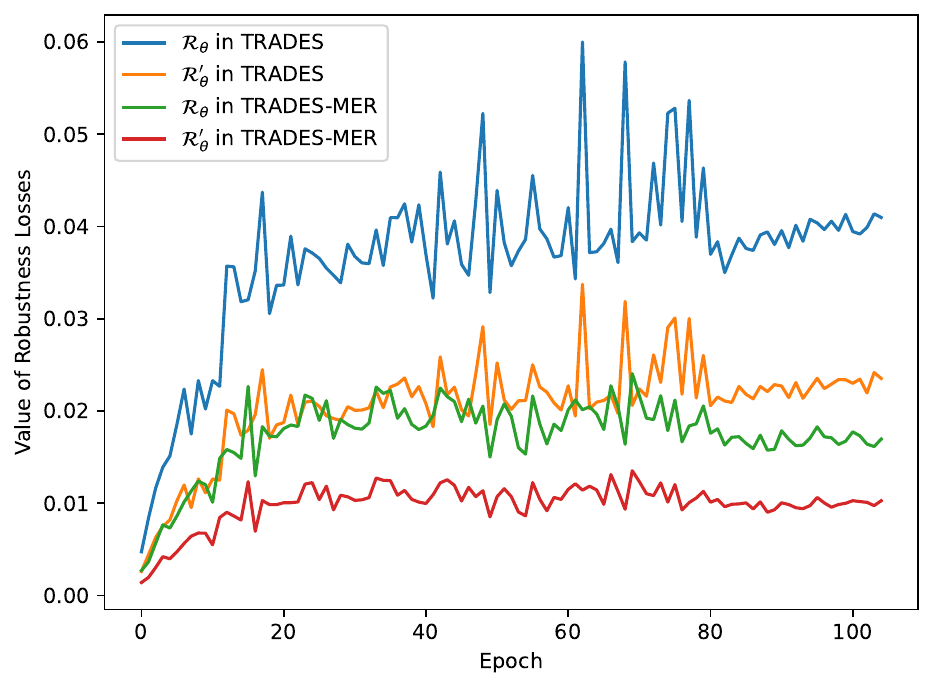}
\caption{Results of $\mathcal{R}_\theta$ and $\mathcal{R}_\theta^\prime$ during the training process of
TRADES and TRADES-MER with the same $\lambda=9$.}
\label{fig:R}
\end{figure}

We first provide the obtained experimental results to support our proposition that FAIT and MER can reduce the difficulty of optimizing $\mathcal{R}_\theta$. In Fig. \ref{fig:R}, we plot the curves of $\mathcal{R}_\theta$ and $\mathcal{R}_\theta^\prime$ for each training epoch of TRADES ($\lambda=9$) and TRADES-MER ($\lambda=9$).
We obtained both $\mathcal{R}_\theta$ and $\mathcal{R}_\theta^\prime$
by averaging the values obtained over a thousand examples in the CIFAR-10 training set \cite{cite47}, and $\mathcal{R}_\theta^\prime$
is calculated with $I=2$.
We can see that
\begin{enumerate}
	\item $\mathcal{R}_\theta^\prime$ is lower than $\mathcal{R}_\theta$ during the training process of both TRADES and TRADES-MER. As $\mathcal{R}_\theta^\prime$ and $\mathcal{R}_\theta$ denote
the robustness loss term of FAIT and TRADES, respectively, this empirical result indicates that the FAIT robustness loss is lower
than that the TRADES robustness loss, which is consistent with the Lemma \ref{lemma:1}.
    \item $\mathcal{R}_\theta$ in TRADES-MER is lower than $\mathcal{R}_\theta$ in TRADES while both TRADES-MER and TRADES train with the
same $\lambda=9$
and this empirical result is consistent with our Theorem \ref{theory:1} and Theorem \ref{theory:2}.
\end{enumerate}
These results demonstrate that FAIT and MER can indeed help reduce the optimization difficulty of $\mathcal{R}_\theta$ in TRADES.

\subsection{Do FAIT and MER enhance the adversarial robustness?}

\subsubsection{Larger $\lambda$ values and better robustness.} Because FAIT and MER can help reduce the difficulty of optimizing the $\mathcal{R}_\theta$, we find that both TRADES-MER and FAIT can thus adopt larger $\lambda$ values than TRADES, and more importantly, when $\lambda$ increases to the optimal
robustness-accuracy tradeoff value, our methods also attain better robustness. In Table \ref{tab:1}, we report the accuracy and AA robustness of TRADES, as well as the FAT for TRADES \cite{cite2}, FAIT and TRADES-MER with different $\lambda$ values on the CIFAR-10 test set with ResNet-18. For FAIT, we use $I=2$, and for TRADES-MER we use \{$\beta_{cle}=1,\beta_{adv}=0$\},
and we provide the parametric search results of $I$ and \{$\beta_{cle},\beta_{adv}$\} in Table \ref{search:1} and Table \ref{search:2},
respectively.

We bold the best AA result of each method. As shown in Table \ref{tab:1}, when TRADES, FAIT and TRADES-MER reach the best robustness at $\lambda=9$, $\lambda=12$ and $\lambda=21$, the robustness of FAIT and TRADES-MER
are both stronger than TRADES.
In addition, we can see that FAIT gains better robustness than FAT for TRADES, which demonstrates that using the 10-step PGD adversary is important for guaranteeing robustness.

\begin{table*}[ht]
\centering
\resizebox{\textwidth}{!}{
	\begin{tabular}{c|cccccccc}
		\hline
		\multirow{2}{*}{$\lambda$} & \multicolumn{2}{c}{TRADES}                                & \multicolumn{2}{c}{FAT for TRADES}                        & \multicolumn{2}{c}{FAIT (ours)}                          & \multicolumn{2}{c}{TRADES-MER (ours)}                     \\
		                           & Clean          & AA                                       & Clean          & AA                                       & Clean          & AA                                      & Clean          & AA                                       \\ \hline
		3                          & 83.62$\pm$0.25 & 46.92$\pm$0.15                           & 84.45$\pm$0.36 & 45.75$\pm$0.14                           & 84.98$\pm$0.81 & 46.84$\pm$1.11                          & 87.14$\pm$0.17 & 44.22$\pm$0.13                           \\
		6                          & 81.45$\pm$0.19 & 48.04$\pm$0.16                           & 82.48$\pm$0.34 & 47.48$\pm$0.23                           & 83.20$\pm$0.17 & 48.82$\pm$0.11                          & 85.64$\pm$0.17 & 46.58$\pm$0.25                           \\
		9                          & 79.42$\pm$0.05 & \textbf{48.53$\pm$0.19} & 81.06$\pm$0.27 & 47.98$\pm$0.11                           & 81.37$\pm$0.06 & 49.40$\pm$0.34                          & 84.69$\pm$0.07 & 47.82$\pm$0.05                           \\
		12                         & 77.91$\pm$0.18 & 48.20$\pm$0.33                           & 80.27$\pm$0.32 & 48.08$\pm$0.32                           & 80.31$\pm$0.16 & \textbf{49.41$\pm$0.1} & 83.52$\pm$0.39 & 48.71$\pm$0.5                            \\
		15                         & 76.60$\pm$0.15 & 48.39$\pm$0.18                           & 79.41$\pm$0.06 & \textbf{48.48$\pm$0.13} & 79.03$\pm$0.09 & 49.15$\pm$0.13                          & 82.54$\pm$0.05 & 48.39$\pm$0.32                           \\
		18                         & 75.74$\pm$0.39 & 47.72$\pm$0.4                            & 78.78$\pm$0.35 & 48.16$\pm$0.15                           & 78.36$\pm$0.09 & 49.05$\pm$0.04                          & 81.58$\pm$0.07 & 48.69$\pm$0.33                           \\
		21                         & 75.03$\pm$0.46 & 47.64$\pm$0.26                           & 78.18$\pm$0.3  & 48.40$\pm$0.19                           & 77.18$\pm$0.3  & 48.72$\pm$0.16                          & 81.23$\pm$0.18 & \textbf{49.04$\pm$0.25} \\ \hline
\end{tabular}
}
\caption{CIFAR-10 results of TRADES, FAT for TRADES, FAIT and TRADES-MER with various $\lambda$.}
\label{tab:1}
\end{table*}
\begin{table}[ht]
	\centering
	\small
	\begin{tabular}{l|lllllllll}
	\hline
$I$   & 1     & 2     & 3     & 4     & 5     & 6     & 7     & 8     & 9     \\ \hline
Clean & 79.29 & 80.31 & 80.57 & 78.99 & 78.32 & 78.80 & 78.03 & 77.73 & 78.30 \\
AA    & 49.07 & \textbf{49.41} & 48.89 & 48.70 & 49.02 & 49.13 & 49.27 & 48.46 & 48.75 \\ \hline
	\end{tabular}
	\caption{The parameter $I$ in FAIT with $\lambda=12$.}
	\label{search:1}
\end{table}
\begin{table}[ht]
\centering
\small
\begin{tabular}{ll|ll}
\hline
$\beta_{cle}$ & $\beta_{adv}$ & Clean & AA    \\ \hline
0             & 1           & 81.14 & 48.9  \\
0.5           & 0.5         & 81.37 & 48.69 \\
1             & 0           & 81.23 & \textbf{49.04} \\ \hline
\end{tabular}
\caption{The parameters $\beta_{cle}$ and $\beta_{adv}$ in TRADES-MER with $\lambda=21$.}
\label{search:2}
\end{table}

\begin{table*}[ht]
\centering
\resizebox{\textwidth}{!}{
\begin{tabular}{c|cccccc}
\hline
\multirow{2}{*}{Method}   & \multicolumn{3}{c}{CIFAR-10}                     & \multicolumn{3}{c}{CIFAR-100}                    \\
                          & Clean          & PGD-100        & AA             & Clean          & PGD-100        & AA             \\ \hline
PGD-AT                    & 83.59$\pm$0.24 & 48.68$\pm$0.15 & 45.83$\pm$0.1  & 58.97$\pm$0.1  & 25.58$\pm$0.39 & 22.79$\pm$0.31 \\
GAIRAT                    & 83.21$\pm$0.23 & 49.30$\pm$0.33 & 46.82$\pm$0.28 & 56.73$\pm$0.25 & 24.78$\pm$0.17 & 22.00$\pm$0.15 \\
MART                      & 78.49$\pm$0.51 & 52.71$\pm$0.18 & 46.24$\pm$0.17 & 53.3$\pm$0.16  & 30.08$\pm$0.26 & 24.61$\pm$0.24 \\
TRADES-AWP                & 81.23$\pm$0.21 & 51.77$\pm$0.06 & 48.62$\pm$0.03 & 57.02$\pm$0.13 & 28.97$\pm$0.33 & 24.56$\pm$0.12 \\
HAT                       & 84.98$\pm$0.14 & 51.61$\pm$0.14 & 48.63$\pm$0.22 & 60.18$\pm$0.33 & 26.84$\pm$0.46 & 22.62$\pm$0.34 \\ \hline
FAIT ($\lambda=12$)       & 80.31$\pm$0.16 & 53.04$\pm$0.08 & 49.41$\pm$0.1  & 57.14$\pm$0.19 & 29.38$\pm$0.03 & 24.79$\pm$0.07 \\
TRADES-MER ($\lambda=21$) & 81.23$\pm$0.18 & 53.4$\pm$0.27  & 49.04$\pm$0.25 & 58.64$\pm$0.08 & 30.04$\pm$0.12 & 24.63$\pm$0.26 \\
FAIT-MER ($\lambda=30$)   & 81.4$\pm$0.28  & \textbf{53.87$\pm$0.24} & \textbf{49.49$\pm$0.37} & 58.87$\pm$0.11 & \textbf{30.48$\pm$0.06} & \textbf{24.84$\pm$0.12} \\ \hline
\end{tabular}
}
\caption{Comparison of the state of the art AT methods with ResNet-18.}
\label{tab:2}
\end{table*}

\subsubsection{Compare with more AT methods.} To further check the effectiveness of the proposed FAIT and MER methods, we compare them with a batch of the state of art AT methods: PGD-AT \cite{cite12}, GAIRAT \cite{cite21}, MART \cite{cite24}, TRADES-AWP \cite{cite20} and HAT \cite{cite19}. We reproduce the learning objectives of PGD-AT, MART and TRADES-AWP in our code, and for GAIRAT and HAT, we use the original code found in their GitHub repositories.

In Table \ref{tab:2},
we report the results obtained on the CIFAR-10 and CIFAR-100 datasets with ResNet-18, and we can see that FAIT and TRADES-MER again outperform the previous works in terms of robustness. In particular, when we combine FAIT with MER (the FAIT-MER), we found that the ResNet-18 model can adopt an enormous $\lambda$ with a value of 30, and FAIT-MER outperforms FAIT and TRADES-MER in robustness to both PGD-100 attacks and AA, which demonstrates the compatibility of FAIT and MER.

\subsection{Scalability}

\subsubsection{Different model architectures.} In Table \ref{tab:3}, we show the CIFAR-10 results obtained by TRADES, FAIT and TRADES-MER under three 
different model architectures: SENET18 \cite{senet}, VGG16 \cite{vgg} and ShuffleNetv2 \cite{shufflenet}. We continue to use the best group of hyperparameters in our ResNet-18 experiments, where $\{\lambda=12,I=2\}$ for FAIT and $\{\lambda=21,\beta_{cle}=1,\beta_{adv}=0\}$ for TRADES-MER. Even though we do not search for the best robustness-accuracy tradeoff hyperparameters for these three architectures, we find that FAIT and TRADES-MER can still outperform TRADES, which demonstrates the scalability of our methods to different model architectures.

\subsubsection{Different $\psi$ functions also work well.}

Recall that the Bregman divergence is parameterized by the convex function $\psi$, as described in Eq. \eqref{eq:bregman_div}; however, our previous analyses of PGD-AT and TRADES were based on $\psi=-H$,
which leaves the following question: \emph{can FAIT and MER still work when
$\psi$ changes to another function?}

Furthermore, we notice that a recent work \cite{cite3} theoretically showed that the KL-divergence in TRADES can be substituted for various statistical distances. Among them, the square error (SE) has been shown to be the most effective one, outperforming the KL-divergence, and this idea is called as the Self-COnsistent Robust Error (SCORE). Because the SE is also a Bregman divergence whose $\psi$ is the squaring function,
$$
    \psi(p) = S(p) = \sum p_{i}^2.
$$
Therefore, we plan to check the effectiveness of FAIT and MER under $\psi=S$. To do so, we first follow
the implementation of SCORE by replacing the KL-divergence in both the inner maximization and outer minimization operations of TRADES with the SE. Then, we expand FAIT and MER to the SE versions.
For FAIT, there is no other difference from Algorithm \ref{alg:fait}. For MER, we maximize $-S$ rather than maximizing the entropy function $H$. We denote the FAIT and MER methods under $\psi=S$ as FAIT$_{\psi=S}$ and TRADES-MER$_{\psi=S}$, respectively.
\begin{table}[t]
    \centering
	 \small
    \begin{tabular}{c|cccccc}
    \hline
    \multirow{2}{*}{Method} & \multicolumn{2}{c}{SENET18} & \multicolumn{2}{c}{VGG16} & \multicolumn{2}{c}{ShuffleNetv2} \\
                            & Clean   & AA              & Clean  & AA             & Clean     & AA                 \\ \hline
    TRADES                  & 81.63     & 49.41           & 77.48    & 43.46          & 71.84       & 38.32              \\
    FAIT                    & 81.70     & 49.60           & 78.21    & \textbf{44.63} & 72.35       & 38.45              \\
    TRADES-MER              & 81.94     & \textbf{49.71}  & 76.97    & 43.60          & 71.79       & \textbf{38.67}     \\ \hline
    \end{tabular}
    \caption{Performance under various model architectures.}
    \label{tab:3}
\end{table}
\begin{table}[t]
    \centering
	\small
    \begin{tabular}{cccccc}
    \hline
    \multirow{2}{*}{Method} & \multirow{2}{*}{$\lambda$} &  & \multicolumn{3}{c}{CIFAR-10} \\
                            &                            &  & Clean   & PGD-100  & AA     \\ \hline
    SCORE                 & 4                          &  & 83.94   & 52.94    & 49.04  \\
    FAIT$_{\psi=S}$         & 8                          &  & 82.11   & 53.93    & \textbf{49.48}  \\
    TRADES-MER$_{\psi=S}$   & 10                         &  & 82.56   & \textbf{54.21}    & 49.42  \\ \hline
    \end{tabular}
        \caption{FAIT and TRADES-MER with $\psi=S$.}
    \label{tab:4}
\end{table}

In Table \ref{tab:4}, we report the CIFAR-10 results of SCORE, FAIT$_{\psi=S}$ and TRADES-MER$_{\psi=S}$ with ResNet-18. For SCORE, we use $\lambda=4$ according to \citeauthor{cite3}, and
we can see that both FAIT$_{\psi=S}$ and TRADES-MER$_{\psi=S}$ can adopt larger $\lambda$ values at the value of 8 and 10, respectively.
Besides, both of them again achieve better robustness, which demonstrates the scalability of our methods when $\psi$ changes.

\section{Conclusion}
\label{sec:6}
In this paper, we demonstrated that reducing the difficulty of optimizing the robustness loss $\mathcal{R}_\theta$
under 10-step PGD adversaries is a promising approach for enhancing adversarial robustness. We build a novel Bregman divergence perspective for AT to clearly look at the optimization problem concerning $\mathcal{R}_\theta$. Based on this perspective, we propose FAIT and MER and verify that both of them can help enhance adversarial robustness and are easier to optimize than their prototype method TRADES. We hope that this novel perspective and our analyses will help future works design more robust DNN models and AT algorithms.

\section{Acknowledgments}
\label*{}
This work was sponsored by Zhejiang Laboratory(No. 2021KD0AB03).

\appendix

\section{Proofs}
\begin{lemma}
\label{Alemma:1}
Given points $x_1$, $x_2$ and $x_*$, if $\exists \alpha \in [0,1]$ such that $p_\theta(x_*) = (1-\alpha) p_\theta(x_1) + \alpha p_\theta(x_2)  $, then the following inequality holds true:
\begin{equation*}
    \begin{aligned}
    KL(p_\theta(x_2) \Vert p_\theta(x_1)) \geq KL(p_\theta(x_2) \Vert p_\theta(x_*)) \\ 
    + KL(p_\theta(x_*) \Vert p_\theta(x_1)).
    \end{aligned}
\end{equation*}
\end{lemma}

\begin{proof}[Proof of Lemma \ref{Alemma:1}]
According to the laws of cosines, we have:

\begin{equation}
    \label{eq:lemma1_proof}
    \begin{aligned}
     & KL(p_\theta(x_2) \Vert p_\theta(x_1)) \\
     &- KL(p_\theta(x_2) \Vert p_\theta(x_*)) - KL(p_\theta(x_*) \Vert p_\theta(x_1) )\\
   = & - \sum (p_\theta(x_2)_i - p_\theta(x_*)_i)(\nabla-H(p_\theta(x_1))_i \\
     & - \nabla-H(p_\theta(x_*))_i) \\
   = & \sum (1-\alpha) \cdot \\
     & \underbrace{(p_\theta(x_2)_i-p_\theta(x_1)_i)}_{\circled{1}} \underbrace{(log(1-\alpha+\alpha \frac{p_\theta(x_2)_i}{p_\theta(x_1)_i}))}_{\circled{2}}.
    \end{aligned}
\end{equation}
Since $\alpha \in [0,1]$, we have $1-\alpha \geq 0$. $\forall x_{1}, x_{2}$, if $p_\theta(x_2)_i > p_\theta(x_1)_i$, we have $\circled{1} > 0 $ and 
$\circled{2} > 0 $, respectively. If $p_\theta(x_2)_i \leq p_\theta(x_1)_i$, we have $\circled{1} \leq 0 $ and 
$\circled{2} \leq 0 $, too. Therefore, Eq. \eqref{eq:lemma1_proof}$\geq 0$, and Lemma 1 holds true.
\end{proof}

\begin{theorem}[$ \mathcal{C}.1$]
\label{theory:1} Given $ \mathcal{H}(f_{\theta_{1}},\epsilon) \leq \mathcal{H}(f_{\theta_{2}},\epsilon)$ and $\mathbb{I}(f_{\theta_{1}},\epsilon) = \mathbb{I}(f_{\theta_{2}},\epsilon)$,
$\forall x \in \mathcal{D}$, if $\mathcal{C}.1$ holds true, then we have $\mathcal{R}_{\theta_{1}}(x,x^\prime) \geq \mathcal{R}_{\theta_{2}}(x,x^\prime) $.
\end{theorem}

\begin{theorem}[$ \mathcal{C}.2 , \mathcal{C}.3$]
\label{theory:2}
Define the difference between the clean and the adversarial probability distribution as $d_{\theta}(x,x^\prime) = p_{\theta}(x) - p_{\theta}(x^\prime) $ 
. Given $ \mathcal{H}(f_{\theta_{1}},\epsilon) \leq \mathcal{H}(f_{\theta_{2}},\epsilon)$ and $\mathbb{I}(f_{\theta_{1}},\epsilon) = \mathbb{I}(f_{\theta_{2}},\epsilon)$,
$\forall x \in \mathcal{D}$,
if $\mathcal{C}.2$ or 
$\mathcal{C}.3$ holds true,
let $d_{\theta_{1}}(x,x^\prime) = d_{\theta_{2}}(x,x^\prime)$, we also have 
$\mathcal{R}_{\theta_{1}}(x,x^\prime) \geq \mathcal{R}_{\theta_{2}}(x,x^\prime) $.
\end{theorem}

\begin{proof}[Proof of Theorem \ref{theory:1}]
For $\mathcal{C}.1$ holds true, model $f_{\theta_1}$ and $f_{\theta_2}$ satisfy
that $ \mathcal{H}(f_{\theta_{1}},\epsilon) \leq \mathcal{H}(f_{\theta_{2}},\epsilon)$ 
and $\mathbb{I}(f_{\theta_{1}},\epsilon) = \mathbb{I}(f_{\theta_{2}},\epsilon)$, we have:
$$
    p_{\theta_1}(x)^T \cdot y \geq p_{\theta_2}(x)^T \cdot y > p_{\theta_2}(x^\prime)^T \cdot y \geq p_{\theta_1}(x^\prime)^T \cdot y
$$
Therefore, in the binary classification case, $\exists \alpha_1,\alpha_2 \in [0,1]$ such that
$$
    p_{\theta_2}(x) = (1-\alpha_1) p_{\theta_1}(x^\prime) + \alpha_1 p_{\theta_1}(x)
$$
$$
    p_{\theta_2}(x^\prime) = (1-\alpha_2) p_{\theta_1}(x^\prime) + \alpha_2 p_{\theta_2}(x)
$$
According to Lemma \ref{Alemma:1}, we have:
\begin{equation}
    \begin{aligned}
    & R_{\theta_1}(x,x^\prime) \\ =  & KL(p_{\theta_1}(x) \Vert p_{\theta_1}(x^\prime)) \\
                            \geq & KL(p_{\theta_1}(x) \Vert p_{\theta_2}(x)) + KL(p_{\theta_2}(x) \Vert p_{\theta_1}(x^\prime)) \\
                            \geq & KL(p_{\theta_1}(x) \Vert p_{\theta_2}(x)) \\ & + KL(p_{\theta_2}(x) \Vert p_{\theta_2}(x^\prime))
                            +KL(p_{\theta_2}(x^\prime) \Vert p_{\theta_1}(x^\prime)) \\
                            = & KL(p_{\theta_1}(x) \Vert p_{\theta_2}(x))  \\
                             & +  KL(p_{\theta_2}(x^\prime) \Vert p_{\theta_1}(x^\prime)) + R_{\theta_2}(x,x^\prime)
    \end{aligned}
\end{equation}
Because KL-divergence is not negative, we have $R_{\theta_1}(x,x^\prime) \geq R_{\theta_2}(x,x^\prime)$, thus Theorem \ref{theory:1} holds true.
\end{proof}

\begin{proof}[Proof of Theorem \ref{theory:2}]
Let $\Delta = d_\theta(x,x^\prime)^T \cdot y $, and because $x^{\prime}$ is more likely to be misclassified than $x$,
we have $\Delta > 0$. Let $t=p_\theta(x^\prime)^T \cdot y$, we have:
\begin{equation*}
\begin{aligned}
      & KL(p_\theta(x) \Vert p_\theta(x^\prime)) \\ = & (t+\Delta) log(\frac{t+\Delta}{t}) + (1-t-\Delta) log(\frac{1-t-\Delta}{1-t}).
\end{aligned}
\end{equation*}

Let $F(t) = -log (\frac{1-t}{t})$, then

\begin{equation}
    \label{eq:kl_t}
\begin{aligned}
      & \frac{\mathrm{d}}{\mathrm{d}t} KL(p_\theta(x) \Vert p_\theta(x^\prime)) \\
    = & log(\frac{t+\Delta}{t}) - \frac{\Delta}{t} - log(\frac{1-t-\Delta}{1-t}) - \frac{\Delta}{1-t} \\
    = & log(\frac{1-t}{t} \cdot \frac{t+\Delta}{1-(t+\Delta)} ) -  \frac{\Delta}{t(1-t)}    \\
    = & F(t+\Delta)-F(t) - \frac{\Delta}{t(1-t)}  \\
    = & F^\prime(\xi)\Delta - \frac{\Delta}{t(1-t)} \\
    = & \frac{\Delta}{\xi(1-\xi)} - \frac{\Delta}{t(1-t)},
\end{aligned}
\end{equation}
where $t<\xi<t+\Delta$ according to the Lagrange's mean value theorem.
For model $f_{\theta_1}$ and $f_{\theta_2}$ satisfy
that $ \mathcal{H}(f_{\theta_{1}},\epsilon) \leq \mathcal{H}(f_{\theta_{2}},\epsilon)$ 
, $\mathbb{I}(f_{\theta_{1}},\epsilon) = \mathbb{I}(f_{\theta_{2}},\epsilon)$ and $d_{\theta_{1}}(x,x^\prime) = d_{\theta_{2}}(x,x^\prime), \forall x \in \mathcal{D}$, we have

1. If $\mathcal{C}.2$ holds true, we have 
$0 < p_{\theta_{1}}(x^\prime)^T \cdot y \leq p_{\theta_{2}}(x^\prime)^T \cdot y \leq \frac{1}{2}$,
and for $0 < t \leq \frac{1}{2}$, we have Eq. \eqref{eq:kl_t}$ < 0 $, therefore
we have $KL(p_{\theta_1}(x) \Vert p_{\theta_1}(x^\prime)) \geq KL(p_{\theta_2}(x) \Vert p_{\theta_2}(x^\prime))$, \emph{i.e.,}
$\mathcal{R}_{\theta_{1}}(x,x^\prime) \geq \mathcal{R}_{\theta_{2}}(x,x^\prime) $.

2. If $\mathcal{C}.3$ holds true, we have 
$\frac{1}{2} < p_{\theta_{2}}(x^\prime)^T \cdot y \leq p_{\theta_{1}}(x^\prime)^T \cdot y < 1$,
and for $\frac{1}{2} < t < 1 $, we have Eq. \eqref{eq:kl_t}$ > 0$, then
we also have
$\mathcal{R}_{\theta_{1}}(x,x^\prime) \geq \mathcal{R}_{\theta_{2}}(x,x^\prime) $. Therefore, Theorem \ref{theory:2} holds true.
\end{proof}
\bibliographystyle{plainnat}
\bibliography{ref}





\end{document}